\documentclass{article}
\usepackage{ijcai19}

\usepackage{times}
\usepackage{soul}
\usepackage{url}
\usepackage[hidelinks]{hyperref}
\usepackage[utf8]{inputenc}
\usepackage[small]{caption}
\usepackage{graphicx}
\usepackage{amsmath}
\usepackage{booktabs}
\title{Towards Fine-grained Large Object Segmentation\\1st Place Solution to 3D AI Challenge 2020 - Instance Segmentation Track}

\author{
Zehui Chen$^{1*}$\and
Qiaofei Li$^{2}$\footnote{The first two authors contributed equally to this work, listed in alphabetical order.}\and
Feng Zhao$^{1}$\\
\affiliations
$^1$NEL-BITA, University of Science and Technology of China\\
$^2$School of Computer Science and Technology, Xidian University\\
\emails
\{lovesnowbest, qjliqiaofei\}@gmail.com,
fzhao956@ustc.edu.cn
}

\begin{document}

\maketitle

\begin{abstract}
This technical report introduces our solutions of Team '\textit{FineGrainedSeg}' for Instance Segmentation track in 3D AI Challenge 2020. In order to handle extremely large objects in 3D-FUTURE, we adopt PointRend as our basic framework, which outputs more fine-grained masks compared to HTC and SOLOv2. Our final submission is an ensemble of 5 PointRend models, which achieves the 1st place on both validation and test leaderboards. The code is available at \href{https://github.com/zehuichen123/3DFuture_ins_seg}{https://github.com/zehuichen123/3DFuture\_ins\_seg}. 
\end{abstract}

\section{Introduction}

\begin{figure}[t]
	\centering\includegraphics[height=1.5in]{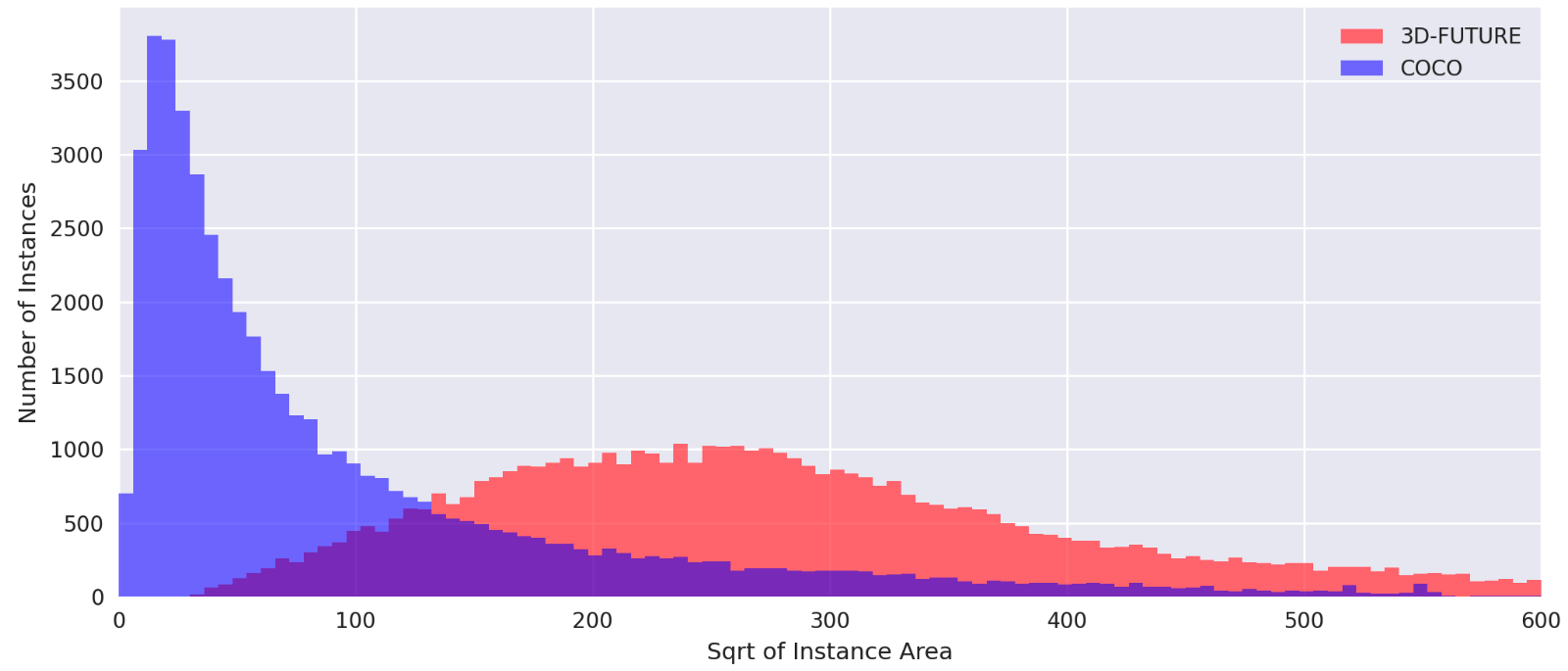}
	\caption{\textbf{Size distribution of bounding boxes in 3D-FUTURE and COCO}. We randomly select 10,000 images for fair comparison. $x$ axis denotes the sqrt area of a bounding box and $y$ axis denotes the number of boxes within each corresponding interval.} 
	\label{fig:scale}
\end{figure}

Deep learning has achieved great success in recent years \cite{fan2019covscode,zhu2019multistep,luo2021rebalancing,luo2023fleet,chen2021disentangle}. Recently, many modern instance segmentation approaches demonstrate outstanding performance on COCO and LVIS, such as HTC~\cite{chen2019hybrid}, SOLOv2~\cite{wang2020solov2}, and PointRend~\cite{kirillov2020pointrend}. Most of these detectors focus on an overall performance on public datasets like COCO, which contains much smaller instances than 3D-FUTURE, while paying less attention to large objects segmentation. As illustrated in Figure \ref{fig:scale}, the size distribution of bounding boxes in 3D-FUTURE and COCO indicates that the former contains much larger objects while the latter is dominated by smaller instances. Thus, the prominent methods used in COCO, like MaskRCNN~\cite{he2017mask} and HTC, may generate blurry contours for large instances. Their mask heads output segmentation from a limited small feature size (e.g., $14\times14$), which is dramatically insufficient to represent large objects. All of these motivate us to segment large instances in a fine-grained and high-quality manner. \\
SOLOv2 builds an efficient single-shot framework with strong performance and dynamically generates predictions with much larger mask size (e.g., 1/4 scale of input size) than HTC. PointRend iteratively renders the output mask over adaptively sampled uncertain points in a coarse-to-fine fashion, which is naturally suitable for generating smooth and fine-grained instance boundaries. By conducting extensive experiments on HTC, SOLOv2 and PointRend, PointRend succeeds in producing finer mask boundaries and significantly outperforms other methods by a large margin. Our step-by-step modifications adopted on PointRend finally achieves state-of-the-art performance on 3D-FUTURE dataset, which yields 79.2 mAP and 77.38 mAP on validation and test set respectively. The final submission is an ensemble of 5 PointRend models with slightly different settings, reaching the 1st place in this competition.


\section{Datasets}

3D-FUTURE dataset is a recently public large-scale indoor dataset 
with 34 categories. Following the official splits, we adopt 12,144 images for training, 2,024 for validation and 6,072 for testing. From the size distribution of bounding boxes in 3D-FUTURE and COCO shown in Figure \ref{fig:scale}, the medium object size of 3D-FUTURE is about 250 while roughly 50 for COCO, indicating that 3D-FUTURE contains much more larger instances\footnote{Followed by 3D-FUTURE official setting, we refer area $\textless 113 \times 113$ for small, $113 \times 113 \sim 256 \times 256$ for medium, and $\textgreater 256 \times 256$ for large, compared to $32 \times 32$ and $96 \times 96$ defined in COCO.}. This distribution divergence motivates us to explore fine-grained large object segmentation methods like PointRend.


\begin{table*}[ht]
\centering
\caption{\textbf{Performance comparison among different methods on validation set (trackA)}. ``DCN" means deformable neural network, ``GC block" means global context block and ``MS Test" means multi-scale testing.}
\begin{tabular}{c|cccccc|ccc|c}
\toprule
Model  & Backbone & DCN & GC block & MaskScoring & MS Test & SyncBN & APs & APm & APl & mAP\\
\hline
HTC    & Res2Net & $\surd$ & $\surd$ & $\surd$ & $\surd$ & $\surd$ & 49.72 & 67.15 & 80.13 & 74.58  \\

SOLOv2 & ResNeXt & $\surd$ & $\surd$ &  & &    & 50.03 & 68.58 & 81.81 & 75.29  \\

PointRend & ResNeXt & $\surd$  &   &    & $\surd$ &  & \textbf{56.23} & \textbf{73.12} & \textbf{85.34} & \textbf{79.17}  \\
\bottomrule
\end{tabular}
\label{tab:plain}
\end{table*}

\begin{table*}[ht]
\centering
\caption{\textbf{PointRend's step-by-step performance on our own validation set (splitted from the original training set)}. ``MP Train" means more points training and ``MP Test" means more points testing. ``P6 Feature" indicates adding P6 to default P2-P5 levels of FPN for both coarse prediction head and fine-grained point head. ``FP16" means mixed precision training.}
\begin{tabular}{c|cccccccc|c}
\toprule
Model & Backbone & Large Resolution & P6 Feature & DCN& MP Train & MP Test & MS Test & FP16 & mAP\\
\hline
MaskRCNN & Res50 &  &  &  &  & & $\surd$ &  & 53.2  \\
\hline
          & Res50 &  &  &  &  & & $\surd$ & & 62.9  \\
          & Res50 &  &  &  &  & $\surd$ & $\surd$ &  & 64.0  \\
PointRend & X101-64x4d &  &  &  &  & $\surd$ & $\surd$ &  & 69.4  \\
          & X101-64x4d & &  & $\surd$ & $\surd$ & $\surd$ & $\surd$ & $\surd$ & 71.6  \\
          & X101-64x4d & $\surd$ & $\surd$ & $\surd$ & $\surd$ & $\surd$ & $\surd$ & $\surd$ & \textbf{74.3}  \\
          
\bottomrule
\end{tabular}
\label{tab:PointRend_val}
\end{table*}

\section{Methods}

In this section, we introduce our practice on three competitive segmentation methods including HTC, SOLOv2 and PointRend. We show step-by-step modifications adopted on PointRend, which achieves better performance and outputs much smoother instance boundaries than other methods.



\subsection{Hybrid Task Cascade}

HTC is known as a competitive method for COCO and OpenImage. By enlarging the RoI size of both box and mask branches to 12 and 32 respectively for all three stages, we gain roughly 4 mAP improvement against the default settings in original paper. Mask scoring head~\cite{huang2019mask} adopted on the third stage gains another 2 mAP. Armed with DCN, GC block and SyncBN training, our HTC with Res2NetR101 backbone yields 74.58 mAP on validation set, as shown in Table \ref{tab:plain}. However, the convolutional mask heads adopted in all stages bring non-negligible computation and memory costs, which constrain the mask resolution and further limit the segmentation quality for large instances.

\subsection{SOLOv2}
Due to limited mask representation of HTC, we move on to SOLOv2, which utilizes much larger mask to segment objects. It builds an efficient yet simple instance segmentation framework, outperforming other segmentation methods like TensorMask~\cite{chen2019tensormask}, CondInst~\cite{tian2020conditional} and BlendMask~\cite{chen2020blendmask} on COCO. In SOLOv2, the unified mask feature branch is dynamically convoluted by learned kernels, and the adaptively generated mask for each location benefits from the whole image view instead of cropped region proposals like HTC. Using ResNeXt101-64x4d plugined with DCN and GC block, SOLOv2 achieves 75.29 mAP on validation set (see Table \ref{tab:plain}). It's worth noting that other attempts, including NASFPN, data augmentation and Mask Scoring, bring little improvement in our experiments.

\begin{table*}[ht]
\centering
\caption{\textbf{PointRend's performance on testing set (trackB)}. ``EnrichFeat" means enhance the feature representation of coarse mask head and point head by increasing the number of fully-connected layers or its hidden sizes. ``BFP" means Balanced Feature Pyramid. Note that BFP and EnrichFeat gain little improvements, we guess that our PointRend baseline already achieves promising performance (77.38 mAP).}
\begin{tabular}{c|cccccc|c|cc|ccc}
\toprule
Model  & Res2Net & ResNeXt & BFP & EnrichFeat & DCN & MS Test & mAP & AP50 & AP75 & APs & APm & APl \\
\hline
           & $\surd$ &   & $\surd$ & $\surd$ & $\surd$ & $\surd$ & 77.21 & \textbf{90.09} & 82.88 & \textbf{47.30} & 71.98 & 81.90 \\
          &  & $\surd$ &  &  & $\surd$ & $\surd$ & \textbf{77.38} & 89.34	& 83.28	& 45.31 & 71.21 & \textbf{82.24} \\
PointRend   &  & $\surd$ & $\surd$ & & $\surd$ & $\surd$ & 77.32 & 89.79	& 83.24	& 45.78 & 72.25 & 81.70 \\
          &  & $\surd$ & $\surd$ & $\surd$ & $\surd$ & $\surd$ & \textbf{77.37} & 89.78	& \textbf{83.39}	& 46.07 & \textbf{72.84} & 81.68 \\
\bottomrule
\end{tabular}
\label{tab:PointRend}
\end{table*}

\subsection{PointRend}\label{section_pointrend}
PointRend performs point-based segmentation at adaptively selected locations and generates high-quality instance mask. It produces smooth object boundaries with much finer details than previously two-stage detectors like MaskRCNN, which naturally benefits large object instances and complex scenes. Furthermore, compared to HTC's mask head, PointRend's lightweight segmentation head alleviates both memory and computation costs dramatically, thus enables larger input image resolutions during training and testing, which further improves the segmentation quality.\\
To fully understand which components contribute to PointRend's performance, we construct our own validation set by randomly selecting 3000 images from original training data to evaluate offline. We will show the step-by-step improvements adopted on PointRend.\\
\textbf{Bells and Whistles.} MaskRCNN-ResNet50 is used as baseline and it achieves 53.2 mAP. For PointRend, we follow the same setting as \cite{kirillov2020pointrend} except for extracting both coarse and fine-grained features from the P2-P5 levels of FPN, rather than only P2 described in the paper. Surprisingly, PointRend yields 62.9 mAP and surpasses MaskRCNN by a remarkable margin of 9.7 mAP. \textbf{More Points Test.} By increasing the number of subdivision points from default 28 to 70 during inference, we gain another 1.1 mAP with free training cost. \textbf{Large Backbone.} X101-64x4d~\cite{xie2017aggregated} is then used as large backbone and it boosts 6 mAP against ResNet50. \textbf{DCN and More Points Train.} We adopt more interpolated points during training, by increasing the number of sampled points from original 14 to 26 for coarse prediction head, and from 14 to 24 for fine-grained point head. Then by adopting DCN~\cite{dai2017deformable}, we gain 71.6 mAP, which already outperforms HTC and SOLOV2 from our offline observation. \textbf{Large Resolution and P6 Feature.} Due to PointRend's lightweight segmentation head and less memory consumption compared to HTC, the input resolution can be further increased from range [800,1000] to [1200,1400] during multi-scale training. P6 level of FPN is also added for both coarse prediction head and fine-grained point head, which finally yields 74.3 mAP on our splitted validation set. Other tricks we tried on PointRend give little improvement, including MaskScoring head, GC Block and DoubleHead~\cite{wu2020rethinking}. \\
In the following, we refer the model in the last row (74.3 mAP) of Table \ref{tab:PointRend_val} as PointRend baseline. The baseline trained on the official training set finally reaches 79.17 and 77.38 mAP on validation and testing set respectively, as shown in Table \ref{tab:plain} and Table \ref{tab:PointRend}. It surpasses SOLOv2 by a large margin: 6.2, 4.5 and 3.5 mAP respectively for small, medium and large size on validation set. We believe that PointRend's iteratively rendering process acts as a pivot for generating high-quality masks, especially fine-grained instance boundaries. Due to its superior performance, we only choose PointRend as ensemble candidates for the final submission. 

\section{Final Submission}
\subsection{Implementation Details}
We implement PointRend using MMDetection~\cite{chen2019mmdetection} and adopt the modifications and tricks mentioned in Section \ref{section_pointrend}. Both X101-64x4d and Res2Net101~\cite{gao2019res2net} are used as our backbones, pretrained on ImageNet only. SGD with momentum 0.9 and weight decay 1e-4 is adopted. The initial learning rate is set to 0.01 for Res2Net101 and 0.02 for X101-64x4d defaultly and decayed by factor 0.1 at epoch 32. During training process, the batch size is 8 (one image per GPU) and all BN statistics are freezed. Mixed precision training enables to reduce GPU memory. The input images are randomly resized to $n \times n$, which is uniformly sampled from range $[1200, 1400]$. All models are trained for 44 epochs. For inference, images are resized to $1400\times1400$ and horizontal flip is used. 

\subsection{Test Performance}
As shown in Table \ref{tab:PointRend}, all PointRend models achieve promising performance. Even without ensemble, our PointRend baseline, which yields 77.38 mAP, has already achieved 1st place on the test leaderboard. Note that several attempts, like BFP~\cite{pang2019libra} and EnrichFeat, give no improvements against PointRend baseline, while they serve as final ensemble candidates. In addition to models listed in Table \ref{tab:PointRend}, another PointRend with slightly different setting (stacking two BFP modules, and increasing the RoIAlign size from original 7 to 10 for bounding box branch) is trained and achieves 76.95 mAP on testing set. So, there are 5 models used for final ensemble.

\subsection{Model Ensemble}
Our final submission is an ensemble of 5 PointRend models. We compare two different ensemble strategies: one is Linear-Reweight~\cite{huang20201st}, and the other is a linear interpolation based on their scores~(Linear-Interpolation). Formally, given a list of model candidates and their scores $S = \{s_1, s_2, ..., s_n\}$, Linear-Interpolation strategy reweights each model with coefficient $w_i$:
\begin{equation}
    w_i = \theta_{min} + \frac{\theta_{max} - \theta_{min}}{max(S) - min(S)}(s_i - min(S)) 
\end{equation}
where $\theta_{min}$ and $\theta_{max}$ are set to 0.6 and 1.0, respectively.\\
To optimize for AP, soft-NMS is adopted. As shown in Table \ref{tab:test_ensemble}, Linear-Interpolation is chosen as final ensemble strategy which boosts the best single model's performance by 1.6 mAP, slightly better than Linear-Reweight. 

\begin{table}[!]
\centering
\caption{\textbf{Final ensemble results on testing set (trackB)}.}
\begin{tabular}{c|c}
\toprule
Method & mAP\\
\hline
Ensemble Candidates & [76.95$\sim$77.38] \\
Linear-Reweight  & 78.92 \\
Linear-Interpolation & \textbf{79.03} \\

\bottomrule
\end{tabular}
\label{tab:test_ensemble}
\end{table}

\begin{figure*}[ht]
\centering\includegraphics[width=1.0\textwidth, height=0.27\textheight]{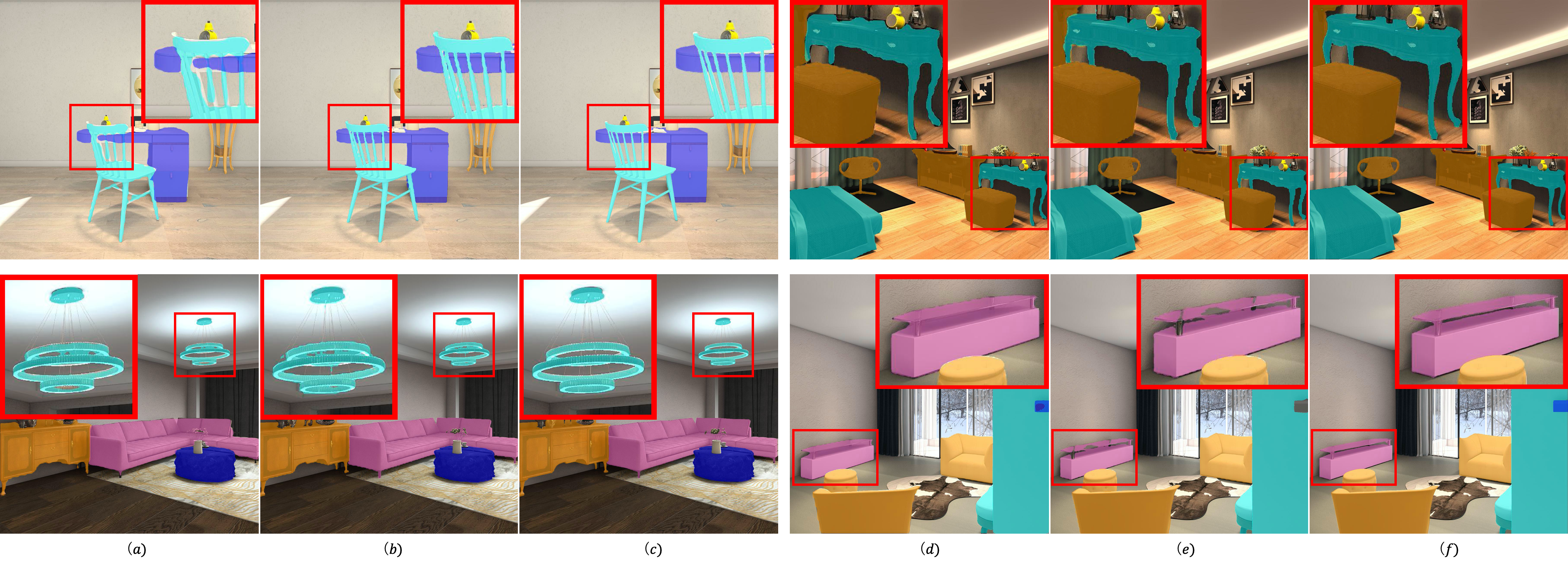}
\captionof{figure}{\textbf{Example of segmentation results on validation dataset from three best single models: (a)(d) HTC, (b)(e) SOLOv2 and (c)(f) PointRend}. PointRend predicts masks with substantially finer details around object boundaries. All figures are best viewed digitally with zoom.}
\label{fig:demo}
\end{figure*}

\section{Visualization}
As shown in Figure~\ref{fig:demo}, we compare HTC, SOLOv2 and PointRend by visualizing their predictions. It can be seen that PointRend generates much finer and smoother segmentation boundaries than HTC and SOLOv2, it also handles overlapped instances gradely (see top-left corner in Figure~\ref{fig:demo}). Meanwhile, PointRend succeeds in distinguishing holes inside objects as background while HTC and SOLOv2 may predict incorrectly as foreground (see bottom line in Figure~\ref{fig:demo}). We attribute PointRend's success to the iteratively rendering process, which performs point-based segmentations at adaptively selected uncertain points and learns to output more fine-grained object contours.

\section{Conclusion}
In this work, we conduct extensive experiments for HTC, SOLOv2 and PointRend on 3D-FUTURE dataset, among which PointRend achieves the best performance and generates smoother object boundaries. By focusing on coarse-to-fine large objects segmentation, our final submission, an ensemble of 5 PointRends, achieves the 1st place for the 3D AI Challenge - Instance Segmentation Track.

\bibliographystyle{named}
\bibliography{ijcai19}
\end{document}